\documentclass[conference]{IEEEtran}
\IEEEoverridecommandlockouts
\usepackage{cite}
\usepackage{amsmath,amssymb,amsfonts}
\usepackage{algorithmic}
\usepackage{graphicx}
\usepackage{subfigure}
\usepackage{tabularx}
\usepackage{array}
\usepackage{booktabs}
\usepackage{textcomp}
\usepackage{xcolor}
\def\BibTeX{{\rm B\kern-.05em{\sc i\kern-.025em b}\kern-.08em
    T\kern-.1667em\lower.7ex\hbox{E}\kern-.125emX}}
\begin{document}

\title{Communication-Efficient Split Learning Based on Analog Communication and Over the Air Aggregation}
\author{\IEEEauthorblockN{Mounssif Krouka, Anis Elgabli, Chaouki ben Issaid, and Mehdi Bennis}
\IEEEauthorblockA{Centre for Wireless Communications (CWC)\\ University of Oulu, Finland\\
Email: \{mounssif.krouka, anis.elgabli, chaouki.benissaid, mehdi.bennis\}@oulu.fi}
}

\maketitle

\begin{abstract}
Split-learning (SL) has recently gained popularity due to its inherent privacy-preserving capabilities and ability to enable collaborative inference for devices with limited computational power. Standard SL algorithms assume an ideal underlying digital communication system and ignore the problem of scarce communication bandwidth. However, for a large number of agents, limited bandwidth resources, and time-varying communication channels, the communication bandwidth can become the bottleneck. To address this challenge, in this work, we propose a novel SL framework to solve the remote inference problem that introduces an additional layer at the agent side and constrains the choices of the weights and the biases to ensure over the air aggregation. Hence, the proposed approach maintains constant communication cost with respect to the number of agents enabling remote inference under limited bandwidth. Numerical results show that our proposed algorithm significantly outperforms the digital implementation in terms of communication-efficiency, especially as the number of agents grows large.
\end{abstract}

\vspace{-0.19in}
\section{Introduction}\vspace{-0.01in}
Deep neural network (DNN) is an efficient and promising approach for solving  machine learning problems, including image classification, speech recognition, and anomaly detection \cite{b1, b2, b3}. However, DNN's performance is often dependent on large model sizes and, as a result, its inference requires high computation complexities, which can become an issue for resource-constrained devices. To overcome this, split-learning (SL), a collaborative inference approach involving communication with a powerful entity, a parameter server (PS), has been proposed in \cite{b4}. 

In SL, NN models are divided on a per-layer basis into an agent-side and server-side segments, respectively. Each agent performs inference using its segment up to a particular layer, known as the \emph{cut layer}. The cut layer outputs are then sent to the PS, which completes the rest of the inference using its segment starting from the \emph{aggregation layer}. By doing so, SL preserves the privacy of the agents' data since they do not share their raw data with the PS, but only the output of the cut layer. Compared to other distributed learning methods, such as federated learning (FL), SL allows a reduction in agent-side computation since the client-side network has fewer layers compared to having the entire network in the FL setting. This is especially important when computations are performed on devices with limited resources \cite{b5}. Furthermore, since the data to be transmitted is limited to the cut layer's output, the agent-side communication costs are significantly reduced. When a larger number of agents is considered, the accuracy of SL is on par with other distributed deep learning methods such as FL and large batch synchronous stochastic gradient descent (SGD), as shown in \cite{b5}. Recently, SL has been proposed to solve communication system problems such as predicting the millimeter-wave received power using camera images and radio frequency signals \cite{b6}, as well as speech command recognition and ECG signal classification in \cite{b7}.

Traditionally, communication systems are assumed to be digital with an ideal communication channel. In digital communication, every agent transmits encoded bits to represent its transmitted output. The PS needs to decode the signal of each agent separately  to ensure error-free decoding. Hence, the agents need to compete on the available communication resources (bandwidth, time). However, as the number of agents grows larger, the communication resources become a bottleneck, negating the practicality of split-learning. 
\begin{figure*}[t]
\centering
\includegraphics[width=16cm,height=12cm,keepaspectratio]{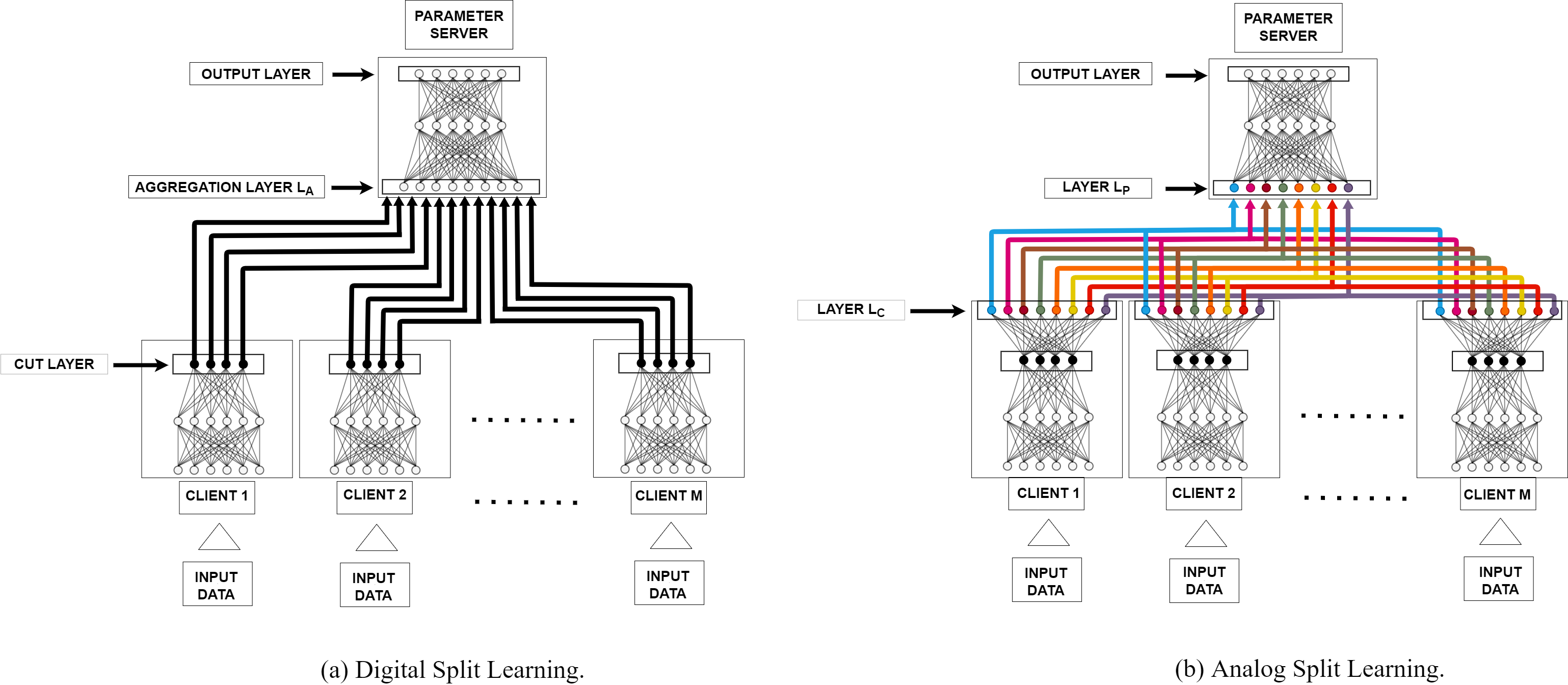}
  \caption{Schematic illustration of the difference between split learning over: (a) digital transmission: outputs are transmitted using orthogonal subcarriers, and (b) analog transmission: agents' outputs are sent through the same subcarrier (same color).}
\label{fig:schema}
\end{figure*} 
Motivated by the fact that the weighted sum of the different agents' outputs is required at every neuron of the first layer of the PS, we propose a novel approach based on analog transmission. Different from the digital scheme, analog over-the-air computation aggregates the agents' transmitted analog signals at each subcarrier. Thus, it requires a number of communication resources that is independent of the number of agents. \\
\textbf{Related works.} Analog schemes have been investigated for FL \cite{b8, b9, b10, b11}.  By leveraging the fact that the wireless MAC automatically provides the PS with the noised version of the aggregate of the gradients, an analog version of the distributed SGD (DSGD) \cite{b8} allows devices to transmit their local gradient estimates directly over the wireless channel. In \cite{b9}, the authors proposed a broadband analog aggregation scheme based on over-the-air computation. Two communication-and-learning trade-offs were formulated to ensure a low-latency FL system. A novel Gradient-Based Multiple Access (GBMA) algorithm was proposed to solve the distributed learning problem over MAC in \cite{b10}, where the nodes use common shaping waveforms to transmit an analog function of the local gradient. The network edge receives and updates the estimate using a superposition of the analog transmitted signals representing a noisy and distorted version of the gradients. Finally, the authors in \cite{b11} proposed A-FADMM, an alternating direction method of multipliers (ADMM)-based approach that only allows the use of the analog transmission's superpositioning property without channel inversion. The FL problem was formulated to account for the impact of the time-varying channel, and ADMM was used to solve the problem while directly integrating the perturbed model updates and protecting the clients' raw data. However, while interesting, these schemes cannot be utilized in SL since the channel inversion yields the sum of the outputs however, what is required is the weighted sum of the outputs between the cut layer and the first layer of the PS, i.e., the aggregation layer.

In this work, we propose a novel approach that cleverly introduces an additional layer at the agent side and constrains the choices of the weights and the biases in this layer and the aggregation layer, to ensure over-the-air aggregation of the analog signals transmitted by different agents at each sub-carrier. With that, the bandwidth is constant with respect to the number of agents and depends only on the number of neurons at the aggregation layer. The details of the proposed approach will be described later. \\
\textbf{Contributions.} Our main contributions are summarized as follows
  
\begin{itemize}
\item To the best of our knowledge, this is the first work proposing analog over the air SL under  noisy wireless channels and limited transmission power.

\item We consider different variants to approximate the cut layer outputs of the agents subject to deep fading channels while benefiting from the analog aggregation and maintaining a good inference accuracy

\item We numerically show that the proposed over-the-air algorithms can adjust to noisy channels and outperform the digital communication counterpart in terms of  completed inference tasks within the available number of channel uses while assuming a large number of agents.
\end{itemize}

The remainder of the paper is organized as follows. The system model is described in Section \ref{sysmodel}. In Section \ref{algo}, we present the proposed analog SL and describe in detail how it adjusts to noisy wireless channels and limited transmission power and how it copes with deep fading channels. Finally, we show some selected numerical results in Section \ref{numsim}.
\vspace{-0.09in}
\section{System Model}\label{sysmodel}
As shown in Fig. \ref{fig:schema}, we consider a remote inference problem where $M$ agents possess input data (e.g. camera images). The input is fed to a local model that consists of multiple NN layers. The output of the cut layer is transmitted to a PS that acquires the outputs of the cut layers from the different agents and feeds them through a number of shared layers to produce a predicted label.

We assume that the split model, distributed between the agents and the PS, has been trained offline, and our goal is  to enable communication-efficient inference over limited bandwidth and noisy and time-variant channels. Instead of relying on \emph{digital communication} where orthogonal subcarriers are allocated to send different elements so that the PS can decode the elements of the agents, we propose a novel approach based on \emph{analog communication} where agents transmit the $i^{th}$ element using the same subcarrier. In the next section, we describe the details of the proposed approach. \vspace{-0.06in}
\section{Proposed Algorithm}\label{algo} \vspace{-0.06in}
Before we dive into explaining the details of the proposed algorithm, we start by determining the communication cost for the digital implementation of SL, as shown in Fig. \ref{fig:schema}(a). Let $C_D$ be the communication cost which is defined as the number of subcarriers needed to transmit the outputs of the cut layer of all agents. For simplicity, we assume that local models of all agents have the same architecture. Using a digital system, the communication cost is given by $C_D = M \cdot N_D$, where $N_D$ is the number of neurons of the cut layer. One drawback of the digital implementation is that the communication cost grows linearly  with respect to the number of agents $M$. 

To overcome this problem, we consider an analog implementation of SL. However, even if we ignore the effect of the wireless channel, direct analog implementation of Fig. \ref{fig:schema}(a) is not possible. The reason is that the aggregation layer does not require the sum of  the outputs of the cut layers but rather a judiciously chosen weighted sum. To solve this issue, we propose a novel implementation of SL over an analog system. As shown in Fig. \ref{fig:schema}(b), the idea is to replace the aggregation layer with two {\it soft layers}:
\begin{itemize}
\item $L_C$ at each agent's side: $L_C$ performs the weight multiplication part as in the original aggregation layer $L_A$. 
\item $L_P$ at the PS' side: $L_P$ performs the bias addition and activation operator part as in the original aggregation layer $L_A$.

\end{itemize} 
Using this approach, we maintain a constant communication cost as the number of agents grows, i.e. $C_A = N_A$, where $N_A$ is the number of neurons at the aggregation layer. 

We will refer to the cut layer $L_C$ and the aggregation layer $L_A$ by layers $l$ and $l+1$, respectively. With that, the output of $j^{th}$ neuron in the $L_C$ layer of the $m^{th}$ agent is given by \vspace{-0.06in}
\begin{align}
O_{l,j}^m = \sum_{k=1}^{K_{l-1}} W_{k,j}^{m}O_{l-1,k}^m, 
\end{align}
where $W_{k,j}^{m}$ is the weight between neurons $k$ and $j$ in the layers $l-1$ and $l$, respectively, at the $m^{th}$ agent and $K_l$ is the number of neurons at layer $l$. Hence, for every neuron in the cut layer, the bias is zero, and the activation function is the identity function. 

To make the explanation easier, we first describe our proposed algorithm while neglecting the effect of the noisy wireless channels, and later in the section, we describe how we adjust to time varying channels. Let us start by writing the input of the $j^{th}$ neuron in layer $L_P$ as \vspace{-0.02in}
\begin{align}\label{input}
I_{l+1,j} = \sum_{m=1}^M O_{l,j}^m. 
\end{align}
Using analog transmission, the $m^{th}$ agent transmits $O_{l,j}^m$ using the $j^{th}$ carrier which is shared across all agents. Hence, only $N_A$ carriers are needed to transmit $\{I_{l+1,j}\}_{j=1}^{N_A}$. Finally, the output of the $j^{th}$ neuron in layer $L_P$ can be written as
\begin{align}
O_{l+1,j} = a_{l+1}(I_{l+1,j} + b_j),
\end{align}
which is equivalent to the output of the original aggregation layer $L_A$ where $a_{l}(\cdot)$ is the activation function at layer $l$.
\vspace{-0.06in}
\subsection{Adaptation to the Noisy Wireless Channels and Limited Transmission Power} One challenge with the implementation of the algorithm is the existence of a noisy time-varying channel between the agents and the PS and the limited transmission power at the agents. Hence, with the assumption of a flat fading channel per subcarrier, if the $m^{th}$ agent sends $x_{m,i}(t)$ over the $i^{th}$ subcarrier  at time $t$, the PS receives
\begin{equation}\label{ch}
y_{m,i}(t) = h_{m,i}(t) \cdot x_{m,i}(t) + n(t),
\end{equation}
where $h_{m,i}(t)$ is the flat fading channel at $i^{th}$ subcarrier between the $m^{th}$ agent and the PS at time $t$, and $n(t)$ is the additive white gaussian (AWGN) noise at the PS at time $t$. Throughout this paper, we assume perfect channel estimation at the transmitter. To ensure that the PS receives $I_{l+1,j}+n(t)$ over the $i^{th}$ subcarrier from all agents, two operations are required to be performed at the agents' side: (i) channel inversion to cancel the effect of the channel, and (ii) power allocation such that the power constraint of each agent is satisfied and equal power across agents is achieved at the PS. Therefore, each agent $m$ needs to calculate its power factor $\alpha_m(t)$ such that $(\alpha_m(t))^2\frac{1}{N_A}\sum^{N_A}_{i=1} |\frac{x_{m,i}(t)}{h_{m,i}}|^2 = P_m$ where $P_m$ is the maximum power. After sending $\alpha_m(t)$ to the PS, the latter determines $\alpha(t) = \min\{\alpha_1(t), \cdots, \alpha_M(t)\}$ and shares it with all the agents over a control channel. Finally, every agent transmits $\alpha(t) \frac{x_{m,i}(t)}{h_{m,i}}$ over the $i^{th}$ subcarrier to the PS. After receiving the aggregated output of all agents, and after matched filtering and dividing by $\alpha(t)$, the PS retrieves $I_{l+1,j}+n(t)$.

\subsection{Adjustment to Deep Fading Channels}\label{Adj_fading}
When $|h_{m,i}(t)|^2 \leq \epsilon$, the $m^{th}$ agent  will not be able to transmit the $i^{th}$ output of its cut layer to the PS. In this case, the PS carries out the inference without the input from agents that are subject to deep fading channels. However, this will have an impact on the quality of the inference. Therefore, we propose to  replace the input of each agent subject to deep fading with the average input across those transmitting at that round. More precisely, if a subset $\mathcal{S}$ of agents, having size $k$, is experiencing deep fading, then the PS considers their output to be \vspace{-0.06in}
\begin{align} 
O^m_{l,j} = \frac{1}{M-k} \sum_{\bar{m} \notin \mathcal{S}}{O^{\bar{m}}_{l,j}}, \,\, \forall \, m \in \mathcal{S}.
\end{align}
This choice might be more suited when there is correlation between the inputs since it is a good estimate of the input of each agent that is not transmitting.

In the remainder, we refer to the implementation of this choice by A-SLv1. When the output of the agents experiencing deep fading is not considered by the PS, we denote the analog implementation in this case by A-SLv0. 

\begin{figure*}[t]
\centering
\includegraphics[width=17.3cm,height=10cm,keepaspectratio]{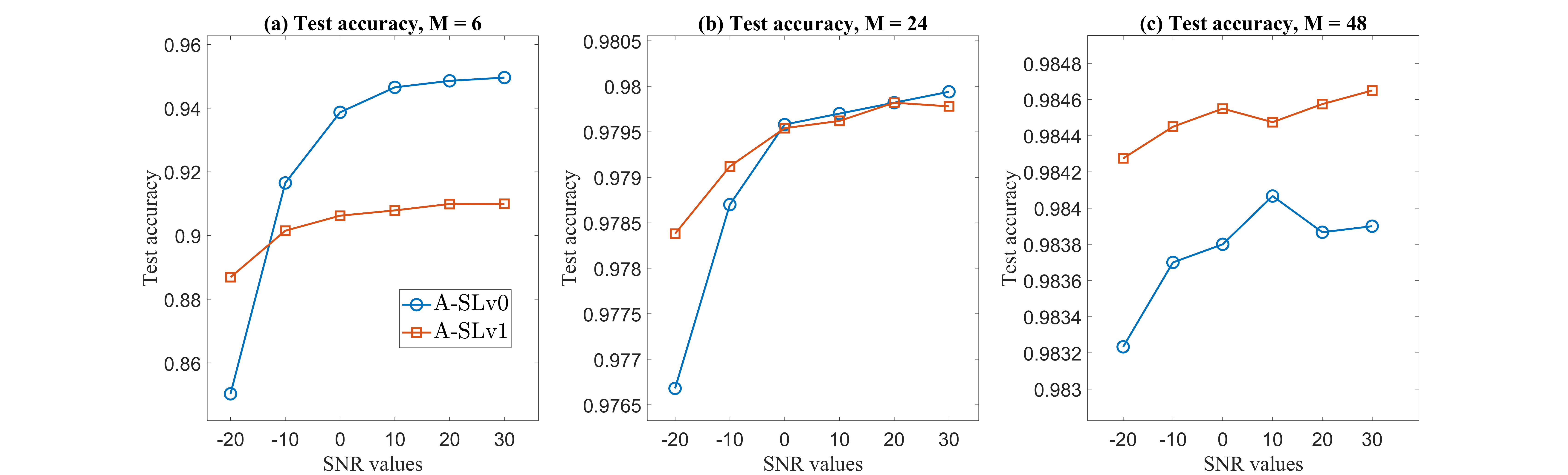}
  \caption{Test accuracy versus SNR for different number of agents. Digital scheme test accuaracy is $99.07\%$, $98.91\%$, and $98.95\%$ for $M =$ $6$, $ 24$, and $48$, respectively.}
\label{final}
\end{figure*}

\section{Numerical Results}\label{numsim}
In this section, we describe the NN architecture, the simulation settings, and the numerical results of our proposed implementations while comparing them to their digital counterpart. \hfill \break
\textbf{Model Architecture.} We consider a NN model that consists of two segements. In the first part, every agent feeds its local data samples to a separate convolutional neural network (CNN). Every CNN consists of 3 convolutional layers (\textit{conv}) with a 3 $\times$ 3 filter and 24, 48, and 72 channels, respectively, followed by a rectified linear unit (\textit{ReLU}) activation function and a \textit{pooling} layer with \textit{stride} 2 for dimensionality reduction. The output of the \textit{conv} layers is fed to a fully-connected layer (\textit{fc}) with $N_D = 32$, then a \textit{ReLU} activation function is applied. The second part is located at the PS side and it consists of 2 \textit{fc} layers: an input layer with $N_A = 256$ neurons followed by a \textit{ReLU} activation function, and an output layer having $10$ neurons where \textit{softmax} function is used. \hfill \break
\textbf{Data Augmentation.}
For the dataset, we consider the MNIST dataset \cite{mnist} which consists of  $28 \times 28$ gray-scale handwritten digits ranging from $0$ to $9$ images with $60K$ for training and $10K$ for testing. In order to supportthe multi-agent scenario, we generated an augmented version of the MNIST dataset using \textit{keras} class \textit{ImageDataGenerator} \cite{keras}. Every agent is fed unique modified versions of the same samples. The samples are augmented using several transformations including rotation, zooming, width and height shifting in order to simulate different camera angles of the same scene. For the training phase, we used a batch size $B=100$, a learning rate $\gamma = 10^{-3}$, and a number of epochs $E=10$.
\begin{figure*}[t]
\centering
\includegraphics[width=16.7cm,height=10cm,keepaspectratio]{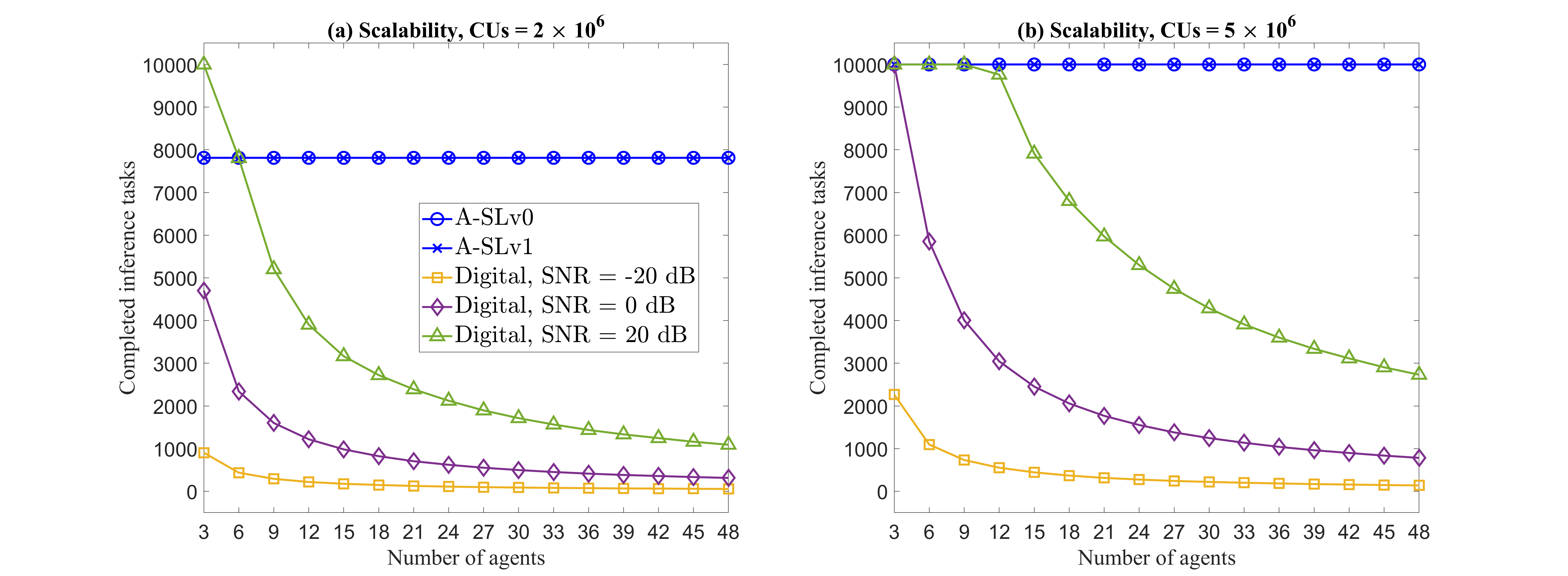}
  \caption{Scalability (number of completed inference tasks w.r.t number of agents) for different $CUs$.}
\label{scalable}
\end{figure*} \hfill \break
\textbf{Network and Communication Environments.}
We simulate our results for different number of agents $M$ when $P_m = 1\,\mathrm{mW}$ and $\epsilon = 0.2$. We consider the number of subcarriers to be $128$. For each plot, we report the mean values based on 5 runs. For analog communication, every $i^{th}$ element of the output of all the agents is transmitted using the same $i^{th}$ subcarrier. In our simulation, the number of transmitted elements is $N_A = 256$, thus the analog scheme requires $\lceil\frac{256}{128}\rceil = 2$ time slots to upload the outputs of all the agents per inference. 
\\
For digital communication, the transmission of every element is carried out using $32$ bits. We note that the number of transmission time slots depends on the number of transmitted elements, as well as the channel gain of each subcarrier. We consider that every subcarrier provides $W_i = 15 \,\mathrm{KHz}$ of bandwidth for a duration of $1\,\mathrm{ms}$ \cite{3gpp}. The channel model is generated following Rayleigh fading with zero mean and unit variance, i.e. $h \sim \mathcal{C N}(0,\,1)$. The required uploading duration $\hat{\tau}_m$ to transmit the output elements of agent $m$ is found as the minimum $\tau_m$ satisfying the following condition:
\vspace{-0.06in}
\begin{align}
\int_{t=0}^{\tau_m} \sum^{\frac{128}{M}}_{i=1}R_{m,i}(t)dt \geq 32N_D,
\end{align}
where $N_0$ is the power spectral density and $R_{m,i}(t) = W_i \log_2\left(1 + {P_m |h_{m,i}(t)|^2}/N_0W_i \right)$. Thus, to complete the inference task, the needed uploading time for all the outputs of the agents is found as $\bar{\tau} = \max\{\hat{\tau}_1, \hat{\tau}_2, \cdots, \hat{\tau}_M\}$. \vspace{-0.07in}
\subsection{Simulation Results}
We show the numerical results in Figs. \ref{final} and \ref{scalable}. In Fig \ref{final}, the performance of the two analog-based algorithms (A-SLv0 and A-SLv1), described in section \ref{Adj_fading}, is reported in terms of the test accuracy with respect to different number of agents and signal to noise ratio (SNR) values. In Fig. \ref{scalable}, we put a constraint on the number of channel uses $CUs = \sum^J_{j=1} S_j$ where $S_j$ is the number of available subcarriers at time slot $j$. Consequently, we plot the number of completed inference tasks (number of full transmissions of the agents' outputs using $10K$ input inference samples) for different number of agents, and we compare the performance of the analog baselines A-SLv0 and A-SLv1 to the digital transmission scheme for different values of SNR = $\{-20,0,20\}$ dB. \hfill \break
\textbf{Impact of Signal to Noise Ratio.}
The average SNR is given by $\mathbb{E}_t\big[ P_m |h_{m,i}(t)|^2/N_0W_i \big] = P_m/N_0W_i$. Consequently, while $N_0 W_i$ is fixed, increasing the SNR is reflected by increasing the transmit power $P_m$. In Fig. \ref{final}, we notice that the test accuracy decreases at low SNR regimes for both analog baselines due to the high effect of the noise incorporated with the received signal at the PS, which affects the inference decision. Nonetheless, the test accuracy improves as we increase the SNR, i.e, the transmit power. In Fig. \ref{scalable}, for the digital scheme, we note that the SNR has a significant effect on the number of completed inference tasks. The better the channel condition (higher SNR) gets, the more inference tasks are completed. \hfill \break
\textbf{Impact of the Number of Agents.} In Fig. \ref{final}(a), we observe that A-SLv0 obtains a better test accuracy compared to A-SLv1. However, the performance of A-SLv1 improves as the number of agents increases. This is justified by the correlation between the input samples, and averaging across the agents which provides useful information to the received signal at the PS improves the inference accuracy. In addition to this, the gap between the test accuracies of the digital scheme and both analog baselines decreases as the number of agents increases. This is mainly due to the fact that the more correlated data is fed to the network, the more robustness and enhancement is instilled into the inference process.
In Fig. \ref{scalable}, as the number of agents increases, the number of completed inference tasks for the digital scheme reduces significantly. This is justified by the fact that the communication cost of the digital scheme grows linearly with the number of agents, and this is limited by the available $CUs$. On the other hand, we note that both A-SLv0 and A-SLv1 have the same fixed number of completed inference tasks. The reason for this is the advantage of analog transmission in view of its communication cost which is solely related to the number of neurons at the aggregation layer. \hfill \break
\textbf{Impact of Constrained amount of Channel Uses.}
In Fig.~\ref{scalable}(a), we fix the number of channel uses to $CUs = 2\times 10^6$. We notice that the number of completed inference tasks using analog transmission is much higher than the number achieved by the digital counterpart, except for $M < 9$, where the communication cost of the digital scheme is less than that of the analog scheme. In Fig. \ref{scalable}(b), the results are shown for a higher number of available channel uses $CUs = 5 \times 10^6$. We notice that both A-SLv0 and A-SLv1 are able to complete all the inference tasks using the available $CUs$. For the digital scheme, more inference tasks are completed compared to the case in Fig. \ref{scalable}(a). However, analog transmission is crucial to fulfil more inference tasks when the number of agents grows and the communication bandwidth is the bottleneck.
\vspace{-0.09in}
\section{Conclusion}\label{conc} \vspace{-0.06in}  
In this paper, we present analog implementations of split learning that adjust to noisy wireless channels and limited transmission power and cope with deep fading channels. Our proposed implementations overcome the drawback of the linear increase in communication cost of the digital communication. Moreover, it enables the weighted sum operation at the aggregation layer by introducing two \textit{soft layers}. Numerical results show that our algorithm is able to cope with noisy and time-varying channels and shows a robust and enhanced performance at high number of agents. Simulations also show that for the same number of available time slots, the number of completed inference tasks is significantly higher when analog transmission is utilized, compared to the digital scheme. \vspace{-0.06in}
\bibliographystyle{IEEEtran}
\bibliography{references}
\end{document}